\documentclass[11pt,a4paper]{article}
\usepackage[hyperref]{emnlp-ijcnlp-2019}
\usepackage{times}
\usepackage{latexsym}

\usepackage{url}

\aclfinalcopy

\usepackage{array}
\usepackage{makecell}
\usepackage{amsmath}
\usepackage{pgfplots}

\usepackage{float}

\usepackage{todonotes}
\usepackage{hyperref}
\usepackage{booktabs}

\makeatletter
\newcommand\footnoteref[1]{\protected@xdef\@thefnmark{\ref{#1}}\@footnotemark}
\makeatother

\usepackage{booktabs}
\usepackage{pgfplots}
\usepackage{amssymb}

\usepackage{algorithm}
\usepackage[noend]{algpseudocode}
\usepackage{scalerel,stackengine,amsmath}
\makeatletter
\def\BState{\State\hskip-\ALG@thistlm}
\makeatother

\newcommand{\tx}[1]{\textit{#1}}
\newcommand{\up}[1]{\textup{#1}}

\title{Analyzing Structures in the Semantic Vector Space: A Framework for Decomposing Word Embeddings}

\author{Andreas Hanselowski$^{\dagger}$$^*$, Iryna Gurevych$^\dagger$$^*$  \\[2mm]
   $^\dagger$Research Training Group AIPHES \\
   \url{https://www.aiphes.tu-darmstadt.de} \\[2mm]
   $^*$Ubiquitous Knowledge Processing Lab (UKP-TUDA)\\
   \url{https://www.ukp.tu-darmstadt.de/}\\[1mm]
   $^\dagger$$^*$ Computer Science Department, Technische Universit\"at Darmstadt}

\date{}

\begin{document}
\maketitle
\begin{abstract}
Word embeddings are rich word representations, which in combination with deep neural networks, 
lead to large performance gains for many NLP tasks. 
However, word embeddings are represented by dense, real-valued vectors 
and they are therefore not directly interpretable. 
Thus, computational operations based on them are also not well understood.
In this paper, we present an approach for analyzing structures in the semantic vector space 
to get a better understanding of the underlying semantic encoding principles.
We present a framework for decomposing word embeddings into smaller meaningful units which we call \emph{sub-vectors}.
The framework opens up a wide range of possibilities analyzing phenomena in vector space semantics, 
as well as solving concrete NLP problems:
We introduce the \emph{category completion task} and show that a sub-vector based approach 
is superior to supervised techniques;
We present a sub-vector based method for solving the \emph{word analogy task},
which substantially outperforms different variants of the traditional \emph{vector-offset method}. 
\end{abstract}

\section{Introduction}

Word embeddings are word representations that are based on the distributional hypothesis \cite{harris1954distributional} 
and express the meaning of a word by a vector.
Due to their expressive power, word embeddings became very popular,
and in combination with deep neural networks, 
significant performance gains have been achieved in many NLP tasks \cite{collobert2011natural, hirschberg2015advances, young2018recent, devlin2018bert}. 
Modern approaches for learning word embeddings are based on the idea of predicting words in a local context window \cite{mikolov2013efficient}.
As a result, low-dimensional vectors are obtained
that capture rich semantic information of a word as has been demonstrated by numerous studies (e.g., \citealt{mikolov2013linguistic}, \citealt{li2015visualizing}).

Nevertheless, since word embeddings are represented by dense, real-valued vectors,
the encoded information cannot be directly interpreted. 
As a result, the computational operations in neural networks based on word embeddings are also not well understood. 
To address the issue and make word embeddings more interpretable, a number of different methods have been proposed:
Rotation of the word embeddings to align the dimension of the vectors with certain attributes \cite{jang2017elucidating, zobnin2017rotations};
transformation of word embeddings into a sparse higher dimensional space where individual dimension represent attributes of word embeddings \cite{C12-1118, fyshe2014interpretable, faruqui2015sparse};
deriving interpretable distributed representations using lexical resources \cite{kocc2018imparting}.

Even though the goal of making word embeddings more transparent is valuable in itself, 
we argue that identifying structures in the semantic vector space is a more beneficial objective.
We believe that this would give us a better theoretical understanding of the underlying semantic encoding principles 
and lead to a more theoretically driven research of distributed representations. 

From this perspective, we consider the following studies as being of particular importance. 
The \emph{vector-offset method} introduced by \citet{mikolov2013linguistic}
allows solving the word analogy task~\cite{jurgens2012semeval} on the basis of offsets between two word vectors.
In fact, the vector-offset analogy is a persistent pattern in the semantic vector space and holds for a broad range of relations \cite{vylomova2015take}.
Follow up work has shown that a word vector can be decomposed into a linear combination of \emph{constituent vectors}
that represent the attributes of this word vector.
\citet{rothe2015autoextend} decomposed word embeddings using WordNet \cite{miller1995wordnet} into representations of lexemes. 
\citet{cotterell2016morphological} presented an approach for deriving representations of morphemes using lexical resources.
These representations can be combined to predict word embeddings for rare words in languages with rich inflectional morphology.
\citet{arora2018linear} used sparse coding to derive 2000 \emph{elementary vectors} from word embeddings called \emph{discourse atoms}.
Going beyond these approaches, in this paper, we further elaborate the assumption that word embeddings are linearly composed from \emph{smaller vectors}.

The contributions of this work are as follows:
(1) We introduce a framework for decomposing word embeddings which does not require lexical resources 
and is not restricted to a fixed set of \emph{elementary vectors}.
Instead, we are able to decompose word vectors into an arbitrary number of vectors
which we call \emph{sub-vectors}, by contrasting different word embeddings whith each other.
The approach is simple and easy to implement, but at the same time very flexible,
as we are able to control which properties we would like to extract from a word vector.
For a rigorous definition of the framework, we introduce the \emph{distributional decomposition hypothesis},
which defines the rules for a legitimate decomposition of a word vector.  

(2) On the basis of the introduced hypothesis, we propose \emph{semantic space networks} (SSNs),
which is an approach for decomposing word embeddings in a systematic manner.
Using SSNs, we analyze semantic and grammatical categories
and are able to identify sub-vectors capturing different attributes of words, 
such as gender, number, and tense.  

(3) We also show that SSNs can be used in a weakly supervised setting.
Given a number of instances of a category, the method allows us to retrieve other words belonging to the same category with high precision. 
We frame this problem setting as the \emph{category completion task} and present two corpora for the problem setting.
We evaluate the performance of our approach on a newly constructed corpus 
and demonstrate that the method is much more data-efficient compared to supervised methods.
Moreover, we present an algorithm to derive sub-vectors for solving the word analogy task \cite{jurgens2012semeval} 
and show that the method outperforms different variants of the \emph{vector offset method} \cite{mikolov2013linguistic}.

The presented framework opens up a wide range of new possibilities for analyzing different phenomena in vector space semantics but also solving concrete NLP problems. 
In this study, we apply the method to traditional word embeddings, such as word2vec \cite{mikolov2013efficient} and GloVe \cite{pennington2014glove}. 
However, the approach is not restricted to \emph{static word embeddings} and can also be used
to analyze \emph{contextualized word embeddings}, such as those derived by ELMo \cite{peters2018deep} and BERT \cite{devlin2018bert}.
Our data and the source code will be made publicly available for future research\footnote{\url{https://github.com/hanselowski/embedding_decomp}}.

\section{Compositionality of word embeddings} \label{sec:hyp}

Modern approaches for learning word embeddings are based on predicting a word given its context or vice versa \cite{mikolov2013efficient}.
The skip-gram model is the most prominent modern approach and
is trained by predicting the context of a word using a neural model. 
It shall serve here as an example model for the following discussion.
Nevertheless, other prediction based models, such as GloVe or Dependency-Based Word Embeddings~\cite{levy2014dependency}, 
are trained in a similar manner and the following discussion also applies to them. 

The goal of the skip-gram model is to maximize the probability  

\vspace{-3mm}
\begin{equation}
 P(u|v) =  \frac{\textup{exp}(u^\intercal  v)}{\sum^n_{i=1}(\textup{exp}(w_i^\intercal v))},
 \label{eq:log}
\end{equation}

where $v$ is the vector of the considered word and $u$ is the vector of a context word~\footnote{In practise, the \emph{Noise Contrastive Estimation} objective function is typically used \cite{mikolov2013distributed}.}.
The vectors $w_i$ represent words, which do not occur in the context of $v$, and serve as negative examples. 
The more frequently a word appears in a particular context, 
the larger the vector gets, which maximizes the probability of this context \cite{schakel2015measuring}.
The context thereby can be defined as a set of words which form a \emph{consistent context window}. 
The context words \emph{mother} and \emph{father}, for example, 
are expected to drive a word vector $v$ in a similar direction, 
whereas \emph{mother} and \emph{vehicle} 
are expected to push the vector to two different points in the semantic space. 
The direction of the resulting vector is therefore associated with its context 
which also defines its meaning. 
The length of the vector, on the other hand, expresses how often it occurs within a particular context window
and it can be therefore interpreted as the \emph{magnitude} of the meaning associated with the context window. 
However, words are often polysemous and carry different meanings.
This implies that they appear in different contexts,
and when the word vector is updated during training, it is simultaneously driven in
different directions.
On the basis of this observation,
\emph{we argue that the resulting word vector, and therefore its meaning, is the sum of the vectors representing its different meanings}.  
If a vector representing a particular meaning of a word is more prominent than its other meanings,
this meaning is stronger represented in the resulting word vector~\footnote{This interpretation is not unique and a similar line of reasoning is presented in \cite{arora2018linear} or \cite{rothe2017autoextend}.}.

Moreover, it must be noted that different contexts are often incompatible, that is, 
the corresponding vectors are pointing in opposing directions in the semantic vector space.
Thus, some of the vector components will cancel each other out and the length of the resulting word vector will be reduced. 
This phenomenon is particularly pronounced for frequent words, 
as these are more often polysemous than less frequent words, as observed by \citet{schakel2015measuring}. 
As a consequence, the resulting word vectors can be considered as \emph{losing a part of the words' meaning.}
We consider the different meanings of a word encoded in a word vector not as discrete separable lexeme vectors \cite{rothe2017autoextend}
but rather as a continuum of an infinite number of \emph{sub-vectors}.  
We believe that the transition between different context windows is not necessarily discrete.
Polysemy, homonymy, metaphor, metonymy, and vagueness \cite{lakoff2008women:2008}
give rise to a continuum of meanings a word can take which extends through the semantic vector space,
rather than forming a number of clearly separable vectors representing the synsets of a word.
Thus, a word vector can be decomposed into an infinite number of sets of sub-vectors,
each of which represents the attributes encoded in a word vector in a different manner. 
Nevertheless, from our perspective, not all possible decompositions of a word-vector are reasonable,
thus, we formally define the properties of a meaningful set of sub-vectors in
the \tx{distributional decomposition hypothesis} below.

\vspace{1mm}
\textbf{Distributional decomposition hypothesis:}

\vspace{1mm}
A word vector $v$ can be linearly decomposed into a finite, \tx{meaningful} set of \tx{sub-vectors}

\vspace{-3mm}
\begin{equation}
 v = \delta_1 + ... + \delta_n, \quad \textup{subv}(v) = [\delta_1, ..., \delta_n]. 
\end{equation}

A vector $\delta$ is considered to be a sub-vector of the word vector $v$,
if the projection of $v$ onto the vector of $\delta$ 
is greater than or equal to the length of the sub-vector $\delta$, i.e.

\vspace{-3mm}
\begin{equation}
\text{subv}(v) = \{\delta \in \mathbb{R}^{n}\, |\, \delta \cdot v\geq \|\delta\|^{2}\}
\end{equation}

We further define the set of word vectors, of which $\delta$ is a sub-vector, as the set of its \emph{children},  i.e. 

\vspace{-3mm}
\begin{equation}
    \text{ch}(\delta) = \left\{v \in \mathbb{R}^{n}\, |\, \delta \in \text{subv}(v) \right\} 
\end{equation}

A sub-vector is considered to be a \emph{meaningful} representation since it is shared by multiple or at least one word vector, which are indicative of its meaning.
More concretely, the meaning of a sub-vector can be derived from the properties that all of its \emph{children} have in common.

\section{Decomposition of word embeddings}

In this section, we introduce a framework for systematically decomposing word embeddings, 
which is in agreement with the distributional decomposition hypothesis.

\subsection{Semantic tree model} \label{sec:semTre}

The semantic tree model allows us to decompose word embeddings but also analyse different linguistic phenomena.  
In this sub-section, we introduce the \emph{semantic tree model} and analyze phenomena like \emph{categorization} and \emph{grammatical categories}.  
Phenomena like \emph{antonymy}, \emph{polysemy}, and \emph{hypernymy} are analyzed in the Appendix~\ref{sec:lex_rel}. 

\subsubsection{Semantic tree model}

A chosen set of word vectors, which are used to define a semantic tree, 
will be referred to as \tx{support vectors} $S = [v_1, ..., v_n]$. 

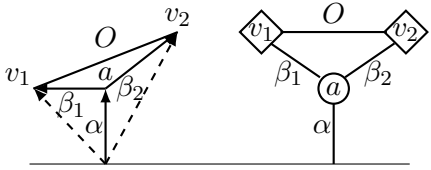
\begin{figure}
\def\x{0.5}
\def\y{0.875}
\def\z{0.5}
\def\a{\z - 0.475}
\def\b{\z + 0.475}
\def\o{0.08}
\def\t{1.5}
\begin{tikzpicture}[scale=2]

    \draw [,] (0.0,0) -- (\z*5,0) ;
    
    \node[] at (\z,\x+0.1) {$a$};
    \node[] at (\z - 0.075,0.25) {$\alpha$};
    \draw [thick, arrows={-latex}] (\z,0) -- (\z,\x);

    \draw [thick, arrows={-latex}] (\z, 0.5) -- (\a,0.5);
    \draw [thick, arrows={-latex}] (\z, 0.5) -- (\b,0.875);

    \node[] at (\a-0.1,\x+\o+0.0) {$v_1$};
    \node[] at (\a+0.25,0.39) {$\beta_1$};
    \draw [thick, dashed, arrows={-latex}] (\z,0) -- (\a,0.505);

    \node[] at (\b,\y+0.1) {$v_2$};
    \node[] at (\b-0.31,0.5) {$\beta_2$};
    \draw [thick, dashed, arrows={-latex}] (\z,0) -- (\b,0.875);

    \node[] at (\z,0.84) {$O$};
    \draw [thick,] (\a, 0.5) -- (\b,0.875);

    \node[] at (\z+\t,\x) {$a$};
    \node[] at (\z - 0.075+\t, 0.25) {$\alpha$};
    \draw [thick,] (\z+\t,0) -- (\z+\t,\x-0.1);
    \draw [thick,] (\z+\t,\x) circle (0.1);

    \draw [thick,] (\z - 0.09 +\t,0.575) -- (\a+0.06+\t,0.8);
    \draw [thick,] (\z + 0.075+\t,0.575) -- (\b-0.075+\t,0.8);

    \node[] at (\a+\t,\y) {$v_1$};
    \node[] at (\a+0.18+\t,0.6) {$\beta_1$};
    \draw[thick,rotate around={45:(\a+\t,\y)}] (\a-0.1+\t,\y-0.1) rectangle (\a+0.1+\t,\y+0.1);

    \node[] at (\b+\t,\y) {$v_2$};
    \node[] at (\b-0.18+\t,0.6) {$\beta_2$};
    \draw[thick,rotate around={45:(\b+\t,\y)}] (\b-0.1+\t,\y-0.1) rectangle (\b+0.1+\t,\y+0.1);

    \node[] at (\z+\t,1) {$O$};
    \draw [thick,] (\a+0.125+\t,0.875) -- (\b-0.125+\t,0.875);
 
\end{tikzpicture}
\centering
\caption{ Semantic tree model in two different representations: (left: vector representation, right: network representation) 
$v_1$ and $v_2$ represented by the dashed lines are the support vectors, $\alpha$ is the root and is the projected vector $v_1$ onto $\alpha_{\textup{unit}}$, $\beta_1$ and $\beta_2$ are the branch sub-vectors, $O$ the vector offset, and $a$ the vector from the origin to the branches, thus $a$ = $\alpha$.
}
\label{fig:tree_1}
\end{figure}

The semantic tree for two word embeddings is schematically illustrated in Figure~\ref{fig:tree_1}. 
The support vectors of the semantic tree in Figure~\ref{fig:tree_1} are the dashed word vectors $v_1$ and $v_2$.

The sub-vector $\alpha$ is defined as the largest vector 
which is shared by all support vectors.
It will be therefore also denoted as the $root$ of the tree.
Since $\alpha$ is shared by all support vectors,
it represents an attribute which all these word vectors have in common.
In order to determine $\alpha$, 
we first need to find its unit vector $\alpha_{\up{unit}} \in \mathbb{R}^n$,
which will be defined as the unit vector of the sum of the support vectors, i.e.

\vspace{-3mm}
\begin{equation}
\alpha_{\up{unit}} = \frac{v_1+v_2+...+v_n}{||v_1 + v_2+...+v_n||}. 
\end{equation}

\citet{mikolov2013distributed} argue that additive composition of word vectors is a good representation of larger text units and link this property to the skip-gram training objective. 
In fact, an average word embeddings representation of a sentence is a 
hard to beat baseline \cite{ruckle2018concatenated}. 
Thus, we use the direction of the word vector sum as the direction of the shared sub-vector.

The sub-vector $\alpha$ is defined as the smallest projected vector onto $\alpha_{\up{unit}}$ of all the support vectors:

\vspace{-3mm}
\begin{equation}
 v_{\up{min}} = \underset{v_i \in S}{\operatorname{min}}( \alpha_{\up{unit}} \cdot v_i ), \quad v_{\up{min}} \in \mathbb{R},
\end{equation}

\vspace{-3mm}
\begin{equation}
 \alpha = v_{\up{min}} \, \alpha_{\up{unit}}, \quad  \alpha \in \mathbb{R}^n.
\end{equation}

%

For the sake of simplicity, 
the operation of deriving the root will be denoted as $\alpha$ = root($S$).
The support vectors do not necessarily correspond to the children of the sub-vector $\alpha$,
as also other word vectors might share $\alpha$, thus $S \subseteq$ ch($\alpha$).
The \tx{branches} of the tree are defined as 
the vectors leading from the top of the root to the individual support vectors.
They are therefore computed by simply subtracting the root from the individual word vectors, i.e. $\beta_{v_i} = v_i - \alpha$.
In Figure~\ref{fig:tree_1}, the two branches are denoted as $\beta_1$ and $\beta_2$.
The branches are also sub-vectors, but in contrast to the root, 
they represent individual attributes of the support vectors.
The procedure of deriving the branches will be denoted as $\beta_i$ = branch$_i$($S$) henceforth.

In addition to the branches, we define the \emph{vector offset} $O$ of the tree.
The vector offset results from the subtraction of one support vector from another
and is identical to the vector offset used to solve the word analogy task by \citet{mikolov2013linguistic}.

The derived semantic tree can be viewed as a \emph{semantic network} 
where the nodes $v_1$, ..., $v_n$ and $a$ represent concepts which are linked by the edges $\beta_1$, ..., $\beta_n$ and $\alpha$.
Whereas the support vectors $v_1$, ...., $v_n$ represent concrete concepts, $a$ is an abstract concept,
which captures the attributes shared by the support vectors $v_1$, ..., $v_n$.
In case we have only two support vectors that are also of equal length,
which for example can be obtained by normalization,
the two branches are of equal length and are pointing exactly in the opposite directions, i.e. 

\vspace{-3mm}
\begin{equation}
||v_1|| = ||v_2||, \quad \beta_{1} = - \beta_{2}.
\end{equation}

This implies that their meanings are opposed to each other.
In such a case, the two branches are also orthogonal to the root, i.e. $\alpha \cdot \beta_{1} = \alpha \cdot \beta_{2} = 0$,
which means that the attributes they describe are unrelated.




The proposed model allows us to separate different properties of word vectors
and therefore, analyze what kind of information is encoded in individual word vectors. 
Thus, in contrast to the cosine similarity, much more nuanced relationships between words can be discovered.
Moreover, the model allows us to identify sub-spaces in the semantic space containing sets of words 
and we are therefore able to perform set-theoretic operations.
In order to illustrate the advantages of the approach,
below, we analyze different lexical relations using the semantic tree model. 
The experimental details are given in the Appendix~\ref{sec:exp}.



\subsubsection{Categorization}

The derivation of the root $\alpha$ of a semantic tree can be viewed as defining the properties of a \emph{category}.
The properties shared by all support vectors are thereby taken as the properties of the category, 
and the set of the children ch($\alpha$), which share these properties, represents the members of the category.

In the Example 1 below, a number of categories are formed using this approach.   
E.g., the sub-vector $\alpha$, which the word vectors $November, December, September, May$ have in common, 
is also shared by eight other word vectors representing the eight remaining months, but no other word vector.
Thus, we are able to form a category on the basis of a couple of examples.
\vspace{3mm}

\noindent \textbf{Example 1.}

\vspace{1mm}
\noindent $S_1$ = [$November, December, September, May$];\\
$\alpha_1$ =  root($S_1$); $|$ch($\alpha_1$)$|$ = 12\\ 
ch$(\alpha_1)$ = [$January, February, March, April,$\\$ May, June, July, August, September, $\\$
October, November, December$]


\vspace{1.5mm}
\noindent $S_2$ = [$hand, foot$];\\
$\alpha_2$ =  root($S_2$); $|$ch($\alpha_2$)$|$ = 9;\\ 
ch$(\alpha_2)$ = [$foot, ankle, wrist, finger, feet, knee,$\\$ 
 elbow, shoulder, hand$];


\vspace{1.5mm}
\noindent $S_3$ = [$man, queen$];\\
$\alpha_3$ =  root($S_3$); $|$ch($\alpha_3$)$|$ = 6;\\ 
ch$(\alpha_3)$= [$ man, king, woman, lady, girl, queen$];

\vspace{2mm}
\noindent $S_4$ = [$car, speed, driver, wheel$];\\
$\alpha_4$ =  root($S_4$); $|$ch($\alpha_4$)$|$ = 24;\\ 
ch$(\alpha_4)$ = [$wheel, driver, car, drivers, SUV, $\\
$ motorcycle, motorists, passenger, crash, ...$];

\subsubsection{Meaning of semantic tree branches}

As discussed in Section~\ref{sec:semTre},
the branches of a semantic tree capture specific meanings of the individual support vectors.
This phenomenon is illustrated in Example 2.
E.g., the branch sub-vector in Example 2 leading from the root to the word vector $Spain$
is also a sub-vector of the word vectors $Barcelona$ and $Madrid$. 
Thus, the words Spain, Barcelona and Madrid are indicative of the meaning of the derived branch sub-vector.
\vspace{1mm}

\noindent \textbf{Example 2.}
\vspace{1mm}\\
\noindent $S$ = [$Spain, France, Russia, Germany, USA$]\\
$\alpha$ =  root($S$); $|$ch($\alpha$)$|$ = 52;\\ 
ch($\alpha$) = [$Germany, France, Spain, USA, $\\
$ Russia, Croatia, Poland, Italy, Serbia, ... $] 

\vspace{1mm}
\noindent ch($\beta_{Spain}) = [Spain, Barcelona, Madrid];$\\
ch($\beta_{France}) = [French, France, Sarkozy];$\\
ch($\beta_{Russia}) = [Russia, Moscow, Putin, $\\$ Kremlin, Ukraine];$\\
ch($\beta_{Germany}) = [Germany, German, Berlin, $\\$  Austria, Frankfurt, Germans];$\\
ch($\beta_{USA}) = [USA];$\\

\subsubsection{Grammatical categories}

In Example 3, we use the semantic tree model in order to identify branch sub-vectors 
which represent grammatical categories.
\vspace{1mm}

\noindent \textbf{Example 3.}
\vspace{1mm}\\
\noindent \textbf{Tense:}\\
\noindent $S$ = [$walk, walked$];\\
ch($\beta_{walk}$) = $[walk];$\\
ch($\beta_{walked}$)= $[ran, struck, walked, crashed, $\\$ threw, drove, stood, sat, fled, grabbed, ...];$
\vspace{2mm}\\
\noindent \textbf{Comperatives, superlatives:}\\
$S$ = [$well, better, best$];\\
ch($\beta_{well}) = [strong];$\\
ch($\beta_{better}) = [better, worse, easier, stronger, $\\$
faster, harder, tougher, weaker, safer, ...];$\\
ch($\beta_{best}) = [best, worst, greatest, fastest, $\\$
strongest, toughest, finest, hottest];$
\vspace{1.5mm}\\
\noindent \textbf{Plural, singular:}\\
$S$ = [$dog, dogs$];\\
ch($\beta_{dog}) = [dog, wallet];$\\
ch($\beta_{dogs})= [dogs, animals, birds, guns, cats, $\\$ planes, pets, horses, prisoners, inmates, ...  ];$

\subsection{Semantic space networks}

The semantic tree model allows decomposing a word vector 
by splitting it up into the two sub-vectors.
However, the derived sub-vectors can be further decomposed 
into more fine-grained representations, 
whereby the two derived sub-vectors serve as support vectors for further semantic trees. 
The derived representations are also sub-vectors but describe more subtle properties compared to the original sub-vectors.
Using this technique, an arbitrary number of trees can be constructed based on the derived sub-vectors in each case. 
The derived trees share sub-vectors and can give rise to networks of arbitrary complexity, which we call \emph{semantic space networks} (SSNs).

In order to illustrate our approach,
in this subsection, we present a constructed \emph{binary tree} as one possible combination of semantic trees.
Further examples can be found in the appendix~\ref{sec:ssn}.

\subsubsection{Binary tree}\label{sec:bintree}

\begin{figure}
\def\x{0.5}
\def\y{0.875}
\def\z{1.0}
\def\a{\z - 0.475}
\def\b{\z + 0.475}
\def\ya{1.25}
\def\za{0.52}
\def\aa{\za - 0.3}
\def\ba{\za + 0.3}
\def\yb{1.25}
\def\zb{1.48}
\def\ab{\zb - 0.3}
\def\bb{\zb + 0.3}
\begin{tikzpicture}[scale=2]

    \draw [,] (0.0,0.1) -- (\z*2,0.1) ;
    
    \node[] at (\z,\x) {$a$};
    \node[] at (\z - 0.09,0.30) {$\alpha$};
    \draw [thick,] (\z,0.1) -- (\z,\x-0.1);
    \draw [thick,] (\z,\x) circle (0.1);

    \draw [thick,] (\z - 0.075, 0.575) -- (\a+0.1,0.8);
    \draw [thick,] (\z + 0.075, 0.575) -- (\b-0.1,0.8);

    \node[] at (\a,\y) {$b$};
    \node[] at (\a+0.2,0.6) {$\beta_1$};
    \draw [thick,] (\a,\y) circle (0.1);

    \node[] at (\b,\y) {$c$};
    \node[] at (\b-0.2,0.6) {$\beta_2$};
    \draw [thick,] (\b,\y) circle (0.1);

    \node[] at (\z,0.98) {$O_1$};
    \draw [thick,] (\a+0.11,0.875) -- (\b-0.11,0.875);

    \draw [thick,] (\za - 0.075, \ya-0.3) -- (\aa+0.075,\ya-0.075);
    \draw [thick,] (\za + 0.075, \ya-0.3) -- (\ba-0.075,\ya-0.075);

    \node[] at (\aa,\ya) {$v_1$};
    \node[] at (\aa+0.55,\ya-0.25) {$\gamma_2$};
    \draw[thick,rotate around={45:(\aa,\ya)}] (\aa-0.1,\ya-0.1) rectangle (\aa+0.1,\ya+0.1);

    \node[] at (\ba,\ya) {$v_2$};
    \node[] at (\ba-0.55,\ya-0.25) {$\gamma_1$};
    \draw[thick,rotate around={45:(\ba,\ya)}] (\ba-0.1,\ya-0.1) rectangle (\ba+0.1,\ya+0.1);

    \node[] at (\za,\ya+0.125) {$O_2$};
    \draw [thick,] (\aa+0.125,\ya) -- (\ba-0.125,\ya);

    \draw [thick,] (\zb - 0.075, \yb-0.3) -- (\ab+0.075,\yb-0.075);
    \draw [thick,] (\zb + 0.075, \yb-0.3) -- (\bb-0.075,\yb-0.075);

    \node[] at (\ab,\yb) {$v_3$};
    \node[] at (\ab+0.55,\yb-0.25) {$\gamma_4$};
    \draw[thick,rotate around={45:(\ab,\yb)}] (\ab-0.1,\yb-0.1) rectangle (\ab+0.1,\yb+0.1);

    \node[] at (\bb,\yb) {$v_4$};
    \node[] at (\bb-0.55,\yb-0.25) {$\gamma_3$};
    \draw[thick,rotate around={45:(\bb,\yb)}] (\bb-0.1,\yb-0.1) rectangle (\bb+0.1,\yb+0.1);

    \node[] at (\zb,\yb+0.125) {$O_3$};
    \draw [thick,] (\ab+0.125,\yb) -- (\bb-0.125,\yb);

\end{tikzpicture}
\centering
\caption{Binary tree network:
Given two pairs of support vectors $[v_1, v_2]$ and $[v_3, v_4]$, two semantic trees can be constructed. Their roots can then be used as support vectors for a third tree giving rise to a binary tree.
}
\label{fig:binTree}
\end{figure}
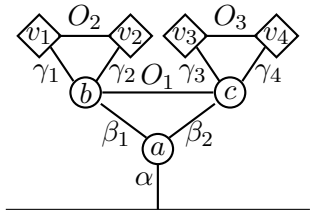

The roots of two semantic trees can serve as support vectors for a third tree, 
which gives rise to a binary tree. 
Such a tree structure is schematically illustrated in Figure~\ref{fig:binTree}.

\vspace{1mm}
\noindent\textbf{Example 4.}\\
\noindent If we choose the four support vectors $S$=$[father, mother, brother, sister]$
for the nodes $v_1$, $v_2$, $v_3$, $v_4$ of the binary tree, 
we obtain the following sub-vectors:\\
ch($\alpha$) = $[brother, daughter, sister, son, mother,$\\$ father, grandmother, grandson, sons, uncle];$\\
ch($b$)  = $[mother, father, daughter, ...];$\\
ch($c$)  = $[brother, sister];$\\
ch($\beta_1$)  = $[woman, child, mother, teacher, baby,$\\$ doctor, abortion, Mrs, babies, Parents, ...];$\\
ch($\beta_2$)  = $[brother, sister, Photo, joins, $\\$ Announces, trademarks, Editing, Join];$\\
ch($\gamma_1$)  = $[father, Sir, legendary, businessman,$\\$  brother,  successor, Sr., grandfather, ...];$\\
ch($\gamma_2$)  = $[mother, she, woman, Sir, girl, baby,$\\$ spokeswoman, herself, Ms., Lady, actress, ...];$\\
ch($\gamma_3$)  = $[brother, brothers, Sir, Gen., $\\$ Councilman, nephew];$\\
ch($\gamma_4$)  = $[sister, spokeswoman, Miss];$\\
\vspace{1mm}
The sub-vector $\alpha$ represents a concept referring to a wide range of different family relations.
The sub-vector represented by the node $b$ ($b = \alpha + \beta_1$) refers to parenthood and the sub-vector $c$ ($c = \alpha + \beta_2$) refers to concepts related to brotherhood and sisterhood.
The sub-vectors $\beta_1$ and $\beta_2$ more specifically refer to these attributes,
whereby the properties represented by $\alpha$ are omitted. 
For $\beta_2$ some noisy terms are introduced. 
The sub-vectors $\gamma_i$ describe more specific attributes mostly associated with gender information.
It must be noted that the tree represents a hierarchy of concepts.
The sub-vector $\alpha$ can be considered as a hypernym of the concepts $b$, $c$, $father$, $mother$, $brother$, and $sister$.
The sub-vector $\beta_1$ can be viewed as a hypernym of the concepts $b$, $father$, $mother$, 
and the sub-vector $\beta_2$ a hypernym of the concepts $c$, $brother$, $sister$.
The, sub-vectors $b$, $c$, can be considered as hypernyms of $father$, $mother,$ and $brother$, $sister$ respectively.
Further examples of SSNs in the appendix~\ref{sec:ssn} demonstrate, 
how sub-vectors can be further decomposed by combining word vectors in different configurations.
Thus, it can be illustrated, which meanings the word vectors have in common and which are distinct.

\section{Experiments}

In this section, we use SSNs to solve a categorization task and the word analogy problem. 
The results of the experiments are discussed in the last sub-section

\subsection{Categorization} \label{sec:cat}

Given a number of concepts, humans are able to construct \emph{ad hoc categories} \cite{barsalou1983ad}, 
which are based on attributes that the given concepts share.
In contrast to formal reasoning systems based on knowledge bases,
which come with a predefined set of categories,
humans are able to form an unlimited number of new categories 
which allows them to solve problems in new situations.
In order to address this problem by machine learning systems,
we define the \emph{category completion task}.
SSNs naturally address the category completion task, 
as the root of a semantic tree defines the attributes which are shared by a number of concepts.
It therefore defines the criteria according to which the remaining members of the category can be found.
As presented in Example 1, the shared sub-vector of the word embeddings $November, December, September, May$
can for instance be used to recover the remaining the eight months which are the other members of the \emph{month category}.

In this section, we perform category completion experiments on two corpora: 
a newly constructed \emph{closed-set category corpus} and
a corpus of categories based on the Google word analogy corpus \cite{mikolov2013efficient}.

\subsubsection{Corpora}

We introduce a closed-set category corpus with 13 categories:
\emph{world\_countries, months, weekdays, digits, rainbow\_colors, planets, family\_relations, personal\_pronouns,  world\_capitals, us\_states,   modals,  possessives, question\_words, }. The number of instances ranges from 7 members as in the category $rainbow$\_colors 
to 116 members as in the category \emph{world\_countries}. The total number of instances is 374.

The Google analogy corpus contains 28 categories with a total number of 1146 instances.
Here closed set categories,  
such as \emph{world\_countries} and \emph{world\_capitals}, are mixed with open set categories,
such as \emph{common\_countries} or sets of adjectives and adverbs. 
The latter cases are difficult to solve since the number of instances is larger than the given sets of words, 
and the category boundaries are fuzzy.

\subsubsection{Categorization experiments}

\begin{table}
\centering
\begin{tabular}{l c c c c    }
  \toprule
  \% data & 10 & 20 & 30 & 40               \\
  \midrule
  baseline           & .182 & .333  & .461 & .571     \\ 
    \midrule
  \multicolumn{3}{l}{\textbf{closed category corpus:}}      &        	     & 	  \\ 
    SSNs              & .349            & .494         & \textbf{.646}          & \textbf{.678} 	     \\
    SVM$_{100}$         & .443            & .435         & .361           & .311          \\ 
    SVM$_{500}$          & \textbf{.582}            & \textbf{.613}         & .585           & .566         \\ 
  \midrule
   \multicolumn{3}{l}{\textbf{Google analogy corpus:}}     &   &     \\ 
    SSNs                & .282    	     & .406         & .329           & .305                    \\ 
    SVM$_{100}$         & .357            & .267         & .218           & .201           \\ 
    SVM$_{500}$          & \textbf{.468}            & \textbf{.474}         & \textbf{.429}           & \textbf{.395}     \\ 
  \bottomrule
\end{tabular}
\caption{F1 scores for SSNs and SVM with different numbers of negative samples (indicated by the subscript number)}
\vspace{-1.5ex}
\label{tb:svm}
\end{table}

To be able to run comparable experiments for categories with a different number of instances,
we provide a certain percentage of instances as example data for each category instead of giving the same number of instances in each case, e.g.: in case we want to perform experiments for 25\% of the data,  we provide two \emph{planet names} out of eight instances in the \emph{planet category} to find the remaining six \emph{planets} or three \emph{month names} from the \emph{month category} to predict the remaining nine.  
In the experiments, we restrict ourselves to a vocabulary of 50,000 to omit rare and noisy word embeddings. 
The defined problem is challenging as the models need to identify a small number of words out of 50,000 instances.\\
In Table \ref{tb:svm}, the performance of the support vector machine (SVM) classifier is compared to the performance of the SSNs on the two corpora using the GloVe word embeddings.
The SVM is superior to other classifiers, such as Logistic Regression, Random Forests, and K-Nearest Neighbors, 
on this task and was therefore chosen as a baseline.
We compute the F1 scores by comparing the example data in combination with the predicted new instances 
to the entire set of instances of the considered category. 
The \emph{baseline} results illustrated in Table~\ref{tb:svm} are obtained by simply considering the example data without predicting any additional instances.
More concretely, we consider the example data as the prediction and take all instances of the category (including the example data) for the evaluation.
To be able to solve the task with an SVM classifier, we consider one category at a time.
We split the entire vocabulary into two classes, 
the instances of the considered category and the remaining words of the vocabulary. 
We then train the SVM on the word vectors of the example data and additional samples from the remaining words in the vocabulary which we call negative samples 
(indicated by the subscript number 100 and 500).
The trained SVM is then used to classify all of the word vectors in the vocabulary 
whether they belong to the considered category or not.
In order to reduce the variance of the results for the SVM and the SSNs, 
we report the mean values of 5 experiment in each of which we randomly exchange the instances in the training and the testing set.
The results in Table \ref{tb:svm} show that compared to the closed category corpus, 
the performance on the Google analogy corpus is significantly lower.
This is due to the problem of open set categories described above.
The results also demonstrate that the SVM requires a large number of negative samples (in addition to example data) in order to reach equivalent performance to SSNs (which only rely on the example data).
SSNs are therefore much more data efficient and require about two orders of magnitude less examples than the SVM model.

\subsection{Word analogy task}\label{sec:ana}

The traditional word analogy task is defined such that 
given two related words from different categories $x^{(1)} \in X$ and $y^{(1)} \in Y$, such as $France$ and $Paris$, and given a third word from the first category $x^{(2)} \in X$, such as $Germany$, one needs to find a word $y^{(2)}$
from the second category $Y$ which is in the same relation to $x^{(2)}$ as $y^{(1)}$ is to $x^{(1)}$.
In the discussed example, \emph{Berlin} satisfies the inferred relation and therefore corresponds to $y^{(2)}$. 
We solve the word analogy task by constructing SSNs and 
extracting sub-vectors which are suitable to predict $y^{(2)}$. 
E.g., to find the vector for \emph{Berlin}, we need to find the sub-vectors representing the abstract concepts of \emph{capital} and \emph{German}. 
To derive \emph{capital}, we can take root sub-vectors from all the members 
in the category capitals $Y$ (e.g., Paris, Rome, Moscow, ...). 
To define \emph{German}, we can take all state names $X$ and take the branch vector for Germany.
This method will be denoted as \emph{SSNbranch}. 
However, branches or individual word vectors contain idiosyncrasies  
which hurt the performance on the word analogy task \cite{drozd2016word}. 
In order to remove idiosyncrasies from the branch sub-vectors, we define Algorithm 1 described below.

\begin{algorithm}
\caption{Filtering algorithm}\label{euclid}
\begin{algorithmic}[1]
\State \textbf{Input}: $x^{(2)}, \, X, \, Y$ where $x^{(2)} \in X$
\State \textbf{Desired output}: $y^{(2)}$
\State for $x_i$ in $X$ not equal to $x^{(2)}$: 
\State  \quad     $\beta_{x^{(2)}_i}$ = branch$_{x^{(2)}}(x^{(2)}, x_i)$
\State root\_$\beta_{x^{(2)}}$ = root($\beta_{x^{(2)}_1}$, ..., $\beta_{x^{(2)}_i}$, ..., $\beta_{x^{(2)}_n}$)
\State root$_Y$ = root($Y$)
\State $\hat{y}^{(2)}$ = root\_$\beta_{x^{(2)}}$ + root$_Y$
\end{algorithmic}
\end{algorithm}

\vspace{-3mm}
The algorithm computes branches $\beta_{x^{(2)}_i}$ for $n-1$ trees (steps 2 and 3). 
Thereby, the word vector $x^{(2)}$ is combined with any other word vector $x_i \in X$ to form trees from which the branches $\beta_{x^{(2)}_i}$ are extracted.
Next, the root of these branches is determined (step 5). 
This branch represents a filtered version of the branch $\beta_{x^{(2)}}$.
In step 6, the root of the category $Y$ is computed.
Finally, an approximation $\hat{y}^{(2)}$ of the word vector $y^{(2)}$ is obtained (step 7).
The resulting method will be denoted as \emph{SSNfilter}. \\
We also compare the results to three variants of the vector-offset method: 
(i) the traditional vector-offset method (\emph{VecOfAdd}) \cite{mikolov2013linguistic}, 
(ii) a definition of the problem as a linear combination of
three pairwise word similarities (\emph{VecOfMul}) \cite{levy2014linguistic}, 
and 
(iii) taking the average offset vector of the given example pairs (\emph{VecOfAvr}) \cite{drozd2016word}, 
The results are illustrated in Table \ref{tb:ana}\footnote{To facilitate reproducibility and comparison of the results we have used the word-embeddings-benchmarks platform in our experiments \url{https://github.com/kudkudak/word-embeddings-benchmarks}}. 
As can be noticed, the vector-offset average method 
\emph{VecOfAvr} is superior to \emph{VecOfAdd} and \emph{VecOfMul}.
Only relying on the original branch sub-vectors using \emph{SSNbranch} yields a worse performance compared to the vector-offset methods.
However, when applying filtering in \emph{SSNfilter}, we are able to substantially outperform the vector-offset methods.
Since the problem setting is deterministic, the results are significant.
Compared to the traditional vector offset methods \emph{VecOfAdd} and \emph{VecOfMul},
\emph{VecOfAvr} and SSNs based methods are using additional information in the form of all the given instances from the categories $X$ and $Y$.
This allows the methods to remove idiosyncrasies and improve performance.

\begin{table}
\centering
\begin{tabular}{l c c  }
  \toprule
  method & GloVe & word2vec   \\
  \midrule
    \emph{VecOfAdd}                       & .717               & .726               \\ 
    \emph{VecOfMul}                       & .725               & .739              \\ 
    \emph{VecOfAvr}                       & .754               & .740      \\ 
    \emph{SSNbranch}                      & .620 	           & .588                 \\ 
    \emph{SSNfilter}                      & \textbf{.797} 	   & \textbf{.781}                \\ 
  \bottomrule
\end{tabular}
\caption{Comparison of different methods on the Google word analogy task}
\vspace{-1.5ex}
\label{tb:ana}
\end{table}

\subsection{Discussion}

The presented experiments show that we can represent various linguistic properties by sub-vectors,
and achieve superior performance compared to other approaches.  
Nevertheless, in our error analysis we have observed that in many cases, 
we cannot perfectly isolate the features of word embeddings.  
There are often words contained in a derived category, which a human would not assign,
(see for instance Example 4: \emph{Photo}, and \emph{joins} are included in the \emph{brother}-\emph{sister} category).
If we use a larger vocabulary of words (a vocabulary of more than 50k words),
even more \emph{foreign} words are included in the derived categories.

On the other hand, the categories almost always contain fewer words than would actually belong to them,
e.g. the categories \emph{comparatives}, \emph{superlatives} and \emph{plural} in Example 3. 
These observations suggest, that the different attributes are not perfectly represented in the semantic vector space,
and the rarer a word is, the less likely it is that it will be contained in the appropriate category.
We suspect that these problems have two different causes: 
(1) The semantic spaces are \emph{irregular} in some regions of the space and are therefore deficient.
(2) The attributes are represented in the semantic vector space but are not linearly separable.
Given the fact that our methods works for a large number of word embeddings,
we believe that in principle, a vector space can be derived where all the attributes are linearly separable,
and this should be explored in future work.



\section{Related Work}

As outlined in the introduction, there has been much work on making word embeddings more interpretable.
Here, we restrict ourselves to a few studies which are most related to the analysis of sub-word representations. 

\citet{yaghoobzadeh2016intrinsic} propose a framework for intrinsic evaluation of word embeddings,
in which they evaluate whether a desired feature is present in a word vector using an SVM classifier. 
\citet{rothe2017autoextend} present a system for learning embeddings for non-word objects like synsets, lexemes, and entities 
for lexical resources such as WordNet. 
\citet{cotterell2016morphological} develop a method for deriving word vectors for rare words 
for languages with rich inflectional morphology.
They rely on morphological resources to derive representations for morphemes which are then linearly combined
to predict a representation of a rare inflection of a word.
Nevertheless, we believe that word vectors possess much more information than can be extracted using lexical resources, and that the proposed decomposition using SSNs allows for a more fine-grained analysis of word embeddings.  
\citet{rothe2016ultradense} introduce a new method for transforming word embeddings into a dense low-dimensional space where the features of interest are represented in each separate dimension.
\cite{arora2018linear} present an approach to derive vectors representing different senses of an ambiguous word. 
They assume that a word vector of a polysemous word is a weighted linear combination of its other meanings,
which is in agreement with our discussion in Section~\ref{sec:hyp}.
Nevertheless, our assumption goes further since we believe that the decomposition of word vectors allows us to analyze the properties encoded in word vectors in general and not just the different senses of an ambiguous word.\\
The vector-offset method introduced by \citet{mikolov2013linguistic}
directly relates to our vector decomposition approach 
as the vector offset also represents specific attributes of word vectors.  
In a number of follow up studies to \citet{mikolov2013linguistic}, different variants of the vector-offset methods have been proposed,
or entirely new approaches presented in order to solve the word analogy task \citet{levy2014linguistic, vylomova2015take, drozd2016word}.
In Section~\ref{sec:ana}, we compare our approach for solving the task to the unsupervised methods.

\section{Conclusion}

In this study, we presented a novel approach for decomposing word embeddings
into meaningful sub-word representations, which we call \emph{sub-vectors}. 
The method allows analyzing the information encoded in a word vector 
or the relation between groups of words. 
For a rigorous definition of the approach, we defined the \emph{distributional decomposition hypothesis}.
On the basis of the defined hypothesis, we introduced \emph{semantic space networks} (SSNs), 
which is a framework for a systematic decomposition of word embeddings. 
Using the proposed framework, we have been able to identify sub-vectors capturing different attributes of words,
such as gender, number or tense.
Moreover, we introduced the \emph{category completion task} and demonstrated that SSNs are much more data efficient 
than supervised classifiers on the task.
We also proposed an approach to solve the word analogy task based on SSNs 
and show that the method outperforms different variants of the vector-offset method.

Important future applications of SSNs lie in diagnostics of models in downstream tasks. 
By decomposing input word embeddings, we can find out what kind of features are feed into a model,
and by adding or removing sub-vectors, we are able manipulate the input. 
We can then analyze, how the changes affect the predictions of the model
and whether it has learned the desired input output relations.


\section{Acknowledgements}

This work has been supported by the German Research Foundation as part of the Research
Training Group ''Adaptive Preparation of Information from Heterogeneous Sources'' (AIPHES) at the Technische Universit\"at Darmstadt under grant No. GRK 1994/1.


\bibliography{emnlp-ijcnlp-2019}

\begin{thebibliography}{33}
\expandafter\ifx\csname natexlab\endcsname\relax\def\natexlab#1{#1}\fi

\bibitem[{Arora et~al.(2018)Arora, Li, Liang, Ma, and
  Risteski}]{arora2018linear}
Sanjeev Arora, Yuanzhi Li, Yingyu Liang, Tengyu Ma, and Andrej Risteski. 2018.
\newblock Linear algebraic structure of word senses, with applications to
  polysemy.
\newblock \emph{Transactions of the Association of Computational Linguistics},
  6:483--495.

\bibitem[{Barsalou(1983)}]{barsalou1983ad}
Lawrence~W Barsalou. 1983.
\newblock Ad hoc categories.
\newblock \emph{Memory \& cognition}, 11(3):211--227.

\bibitem[{Collobert et~al.(2011)Collobert, Weston, Bottou, Karlen, Kavukcuoglu,
  and Kuksa}]{collobert2011natural}
Ronan Collobert, Jason Weston, L{\'e}on Bottou, Michael Karlen, Koray
  Kavukcuoglu, and Pavel Kuksa. 2011.
\newblock Natural language processing (almost) from scratch.
\newblock \emph{Journal of machine learning research}, 12(Aug):2493--2537.

\bibitem[{Cotterell et~al.(2016)Cotterell, Sch{\"u}tze, and
  Eisner}]{cotterell2016morphological}
Ryan Cotterell, Hinrich Sch{\"u}tze, and Jason Eisner. 2016.
\newblock Morphological smoothing and extrapolation of word embeddings.
\newblock In \emph{Proceedings of the 54th Annual Meeting of the Association
  for Computational Linguistics (Volume 1: Long Papers)}, volume~1, pages
  1651--1660.

\bibitem[{Devlin et~al.(2018)Devlin, Chang, Lee, and
  Toutanova}]{devlin2018bert}
Jacob Devlin, Ming-Wei Chang, Kenton Lee, and Kristina Toutanova. 2018.
\newblock Bert: Pre-training of deep bidirectional transformers for language
  understanding.
\newblock \emph{arXiv preprint arXiv:1810.04805}.

\bibitem[{Drozd et~al.(2016)Drozd, Gladkova, and Matsuoka}]{drozd2016word}
Aleksandr Drozd, Anna Gladkova, and Satoshi Matsuoka. 2016.
\newblock Word embeddings, analogies, and machine learning: Beyond king-man+
  woman= queen.
\newblock In \emph{Proceedings of COLING 2016, the 26th International
  Conference on Computational Linguistics: Technical Papers}, pages 3519--3530.

\bibitem[{Faruqui et~al.(2015)Faruqui, Tsvetkov, Yogatama, Dyer, and
  Smith}]{faruqui2015sparse}
Manaal Faruqui, Yulia Tsvetkov, Dani Yogatama, Chris Dyer, and Noah Smith.
  2015.
\newblock Sparse overcomplete word vector representations.
\newblock \emph{arXiv preprint arXiv:1506.02004}.

\bibitem[{Fyshe et~al.(2014)Fyshe, Talukdar, Murphy, and
  Mitchell}]{fyshe2014interpretable}
Alona Fyshe, Partha~P Talukdar, Brian Murphy, and Tom~M Mitchell. 2014.
\newblock Interpretable semantic vectors from a joint model of brain-and
  text-based meaning.
\newblock In \emph{Proceedings of the conference. Association for Computational
  Linguistics. Meeting}, volume 2014, page 489. NIH Public Access.

\bibitem[{Harris(1954)}]{harris1954distributional}
Zellig~S Harris. 1954.
\newblock Distributional structure.
\newblock \emph{Word}, 10(2-3):146--162.

\bibitem[{Hirschberg and Manning(2015)}]{hirschberg2015advances}
Julia Hirschberg and Christopher~D Manning. 2015.
\newblock Advances in natural language processing.
\newblock \emph{Science}, 349(6245):261--266.

\bibitem[{Jang and Myaeng(2017)}]{jang2017elucidating}
Kyoung-Rok Jang and Sung-Hyon Myaeng. 2017.
\newblock Elucidating conceptual properties from word embeddings.
\newblock In \emph{Proceedings of the 1st Workshop on Sense, Concept and Entity
  Representations and their Applications}, pages 91--95.

\bibitem[{Jurgens et~al.(2012)Jurgens, Turney, Mohammad, and
  Holyoak}]{jurgens2012semeval}
David~A Jurgens, Peter~D Turney, Saif~M Mohammad, and Keith~J Holyoak. 2012.
\newblock Semeval-2012 task 2: Measuring degrees of relational similarity.
\newblock In \emph{Proceedings of the First Joint Conference on Lexical and
  Computational Semantics-Volume 1: Proceedings of the main conference and the
  shared task, and Volume 2: Proceedings of the Sixth International Workshop on
  Semantic Evaluation}, pages 356--364. Association for Computational
  Linguistics.

\bibitem[{Ko{\c{c}} et~al.(2018)Ko{\c{c}}, Utlu, Senel, and
  Ozaktas}]{kocc2018imparting}
Aykut Ko{\c{c}}, Ihsan Utlu, Lutfi~Kerem Senel, and Haldun~M Ozaktas. 2018.
\newblock Imparting interpretability to word embeddings.
\newblock \emph{arXiv preprint arXiv:1807.07279}.

\bibitem[{Lakoff(2008)}]{lakoff2008women:2008}
George Lakoff. 2008.
\newblock Women, fire, and dangerous things.
\newblock University of Chicago press.

\bibitem[{Levy and Goldberg(2014{\natexlab{a}})}]{levy2014dependency}
Omer Levy and Yoav Goldberg. 2014{\natexlab{a}}.
\newblock Dependency-based word embeddings.
\newblock In \emph{Proceedings of the 52nd Annual Meeting of the Association
  for Computational Linguistics (Volume 2: Short Papers)}, volume~2, pages
  302--308.

\bibitem[{Levy and Goldberg(2014{\natexlab{b}})}]{levy2014linguistic}
Omer Levy and Yoav Goldberg. 2014{\natexlab{b}}.
\newblock Linguistic regularities in sparse and explicit word representations.
\newblock In \emph{Proceedings of the eighteenth conference on computational
  natural language learning}, pages 171--180.

\bibitem[{Li et~al.(2015)Li, Chen, Hovy, and Jurafsky}]{li2015visualizing}
Jiwei Li, Xinlei Chen, Eduard Hovy, and Dan Jurafsky. 2015.
\newblock Visualizing and understanding neural models in nlp.
\newblock \emph{arXiv preprint arXiv:1506.01066}.

\bibitem[{Mikolov et~al.(2013{\natexlab{a}})Mikolov, Chen, Corrado, and
  Dean}]{mikolov2013efficient}
Tomas Mikolov, Kai Chen, Greg Corrado, and Jeffrey Dean. 2013{\natexlab{a}}.
\newblock Efficient estimation of word representations in vector space.
\newblock \emph{arXiv preprint arXiv:1301.3781}.

\bibitem[{Mikolov et~al.(2013{\natexlab{b}})Mikolov, Sutskever, Chen, Corrado,
  and Dean}]{mikolov2013distributed}
Tomas Mikolov, Ilya Sutskever, Kai Chen, Greg~S Corrado, and Jeff Dean.
  2013{\natexlab{b}}.
\newblock Distributed representations of words and phrases and their
  compositionality.
\newblock In \emph{Advances in neural information processing systems}, pages
  3111--3119.

\bibitem[{Mikolov et~al.(2013{\natexlab{c}})Mikolov, Yih, and
  Zweig}]{mikolov2013linguistic}
Tomas Mikolov, Wen-tau Yih, and Geoffrey Zweig. 2013{\natexlab{c}}.
\newblock Linguistic regularities in continuous space word representations.
\newblock In \emph{Proceedings of the 2013 Conference of the North American
  Chapter of the Association for Computational Linguistics: Human Language
  Technologies}, pages 746--751.

\bibitem[{Miller(1995)}]{miller1995wordnet}
George~A Miller. 1995.
\newblock Wordnet: a lexical database for english.
\newblock \emph{Communications of the ACM}, 38(11):39--41.

\bibitem[{Murphy et~al.(2012)Murphy, Talukdar, and Mitchell}]{C12-1118}
Brian Murphy, Partha Talukdar, and Tom Mitchell. 2012.
\newblock Learning effective and interpretable semantic models using
  non-negative sparse embedding.
\newblock In \emph{Proceedings of COLING 2012}, pages 1933--1950. The COLING
  2012 Organizing Committee.

\bibitem[{Pennington et~al.(2014)Pennington, Socher, and
  Manning}]{pennington2014glove}
Jeffrey Pennington, Richard Socher, and Christopher~D. Manning. 2014.
\newblock \href {http://www.aclweb.org/anthology/D14-1162} {Glove: Global
  vectors for word representation}.
\newblock In \emph{Empirical Methods in Natural Language Processing (EMNLP)},
  pages 1532--1543.

\bibitem[{Peters et~al.(2018)Peters, Neumann, Iyyer, Gardner, Clark, Lee, and
  Zettlemoyer}]{peters2018deep}
Matthew~E Peters, Mark Neumann, Mohit Iyyer, Matt Gardner, Christopher Clark,
  Kenton Lee, and Luke Zettlemoyer. 2018.
\newblock Deep contextualized word representations.
\newblock \emph{arXiv preprint arXiv:1802.05365}.

\bibitem[{Rothe et~al.(2016)Rothe, Ebert, and
  Sch{\"u}tze}]{rothe2016ultradense}
Sascha Rothe, Sebastian Ebert, and Hinrich Sch{\"u}tze. 2016.
\newblock Ultradense word embeddings by orthogonal transformation.
\newblock \emph{arXiv preprint arXiv:1602.07572}.

\bibitem[{Rothe and Sch{\"u}tze(2015)}]{rothe2015autoextend}
Sascha Rothe and Hinrich Sch{\"u}tze. 2015.
\newblock Autoextend: Extending word embeddings to embeddings for synsets and
  lexemes.
\newblock \emph{arXiv preprint arXiv:1507.01127}.

\bibitem[{Rothe and Sch{\"u}tze(2017)}]{rothe2017autoextend}
Sascha Rothe and Hinrich Sch{\"u}tze. 2017.
\newblock Autoextend: Combining word embeddings with semantic resources.
\newblock \emph{Computational Linguistics}, 43(3):593--617.

\bibitem[{R{\"u}ckl{\'e} et~al.(2018)R{\"u}ckl{\'e}, Eger, Peyrard, and
  Gurevych}]{ruckle2018concatenated}
Andreas R{\"u}ckl{\'e}, Steffen Eger, Maxime Peyrard, and Iryna Gurevych. 2018.
\newblock Concatenated $ p $-mean word embeddings as universal cross-lingual
  sentence representations.
\newblock \emph{arXiv preprint arXiv:1803.01400}.

\bibitem[{Schakel and Wilson(2015)}]{schakel2015measuring}
Adriaan~MJ Schakel and Benjamin~J Wilson. 2015.
\newblock Measuring word significance using distributed representations of
  words.
\newblock \emph{arXiv preprint arXiv:1508.02297}.

\bibitem[{Vylomova et~al.(2015)Vylomova, Rimell, Cohn, and
  Baldwin}]{vylomova2015take}
Ekaterina Vylomova, Laura Rimell, Trevor Cohn, and Timothy Baldwin. 2015.
\newblock Take and took, gaggle and goose, book and read: Evaluating the
  utility of vector differences for lexical relation learning.
\newblock \emph{arXiv preprint arXiv:1509.01692}.

\bibitem[{Yaghoobzadeh and Sch{\"u}tze(2016)}]{yaghoobzadeh2016intrinsic}
Yadollah Yaghoobzadeh and Hinrich Sch{\"u}tze. 2016.
\newblock Intrinsic subspace evaluation of word embedding representations.
\newblock \emph{arXiv preprint arXiv:1606.07902}.

\bibitem[{Young et~al.(2018)Young, Hazarika, Poria, and
  Cambria}]{young2018recent}
Tom Young, Devamanyu Hazarika, Soujanya Poria, and Erik Cambria. 2018.
\newblock Recent trends in deep learning based natural language processing.
\newblock \emph{IEEE Computational intelligenCe magazine}, 13(3):55--75.

\bibitem[{Zobnin(2017)}]{zobnin2017rotations}
Alexey Zobnin. 2017.
\newblock Rotations and interpretability of word embeddings: The case of the
  {Russian} language.
\newblock In \emph{International Conference on Analysis of Images, Social
  Networks and Texts}, pages 116--128. Springer.

\end{thebibliography}
\bibliographystyle{acl_natbib}


\appendix


\section{Appendix}

\subsection{Analysis of lexical relations: experimental details}\label{sec:exp}

The experiments have been performed using pretrained, unnormalized word2vec embeddings\footnote{\label{fn:w2v}https://code.google.com/archive/p/word2vec/}.
Since the vector space is more regular for more frequent words, 
we restrict the vocabulary to 11,000 highest ranked words in the word2vec vocabulary.
In order to further reduce the influence of \emph{noisy} word-vectors, we have omitted all multi-word expressions.

\subsection{Analysis of lexical relation using the semantic tree model} 
\label{sec:lex_rel}

\subsubsection{Antonymy}

The evaluation of the antonymy relation using word embeddings is problematic
as the antonym word vectors are often not symmetric.
In in the antonymy relation analyzed in Example 5,
the word vector for \emph{woman}, for example, is significantly larger than the vector for \emph{man}.
This is because the word \emph{man} is used in more contexts and has therefore more meanings. 
As discussed in Section~\ref{sec:hyp}, 
the word vector in such cases ''loses'' meaning and becomes shorter.
As a result, the branch vector for \emph{man} is shorter than the one for \emph{woman}. 
Moreover, the branch vector for \emph{woman} has much more children word vectors,
which means that it captures richer semantic information. 
However, it also includes words, which are in general not associated with the attribute $female$, such as $child$ or $husband$. 
We therefore also derive the orthogonal component vector of the vector $\beta_{woman}$ 
to the root, which we denote by $\beta^\perp_{woman}$.
It must be noted that this vector is in exact opposition to the branch vector $\beta_{man}$. 
The branch vector $\beta^\perp_{woman}$ has less children vectors but these are more obviously associated to the attribute of being $female$.
\vspace{4mm}\\
\noindent \textbf{Example 5.}
\vspace{1mm}\\
\noindent $s$ = [$man, woman$];\\
$||man||$ = 2.311; $||woman||$ = 2.656;\\
$\alpha$ =  root$(s)$; $|$ch($\alpha$)$|$ = 2;\\  
ch($\alpha$) = [$man, woman$];

\vspace{2mm}
\noindent ch($\beta_{man}$)  = $[man, guys, guy, Man];$

\vspace{2mm}
\noindent  $|$ch($\beta_{woman}$)$|$  = 46; \\ 
ch($\beta_{woman}$) = $[her, she, She, women, $\\$  woman, child, mother, daughter, husband, ...];$

\vspace{2mm}
\noindent $|$ch($\beta^\perp_{woman}$)$|$  = 17 \\ 
ch($\beta^\perp_{woman}$) = $[actress, herself, $\\
$ pregnancy,   spokeswoman,   Ms., Women,  ...];$

\subsubsection{Polysemy}

Polysemous word have a number of different meanings,
which are to some extend represented in their word embeddings.
Using the semantic space tree model, 
we can to some degree recover a vector which represents a particular \emph{synset} of the considered word.
As illustrated in Example 6, 
by subtracting the meaning of different words associated with \emph{chairman} from the word vector of \emph{chair},
one can to some extend recover its second meaning, namely that of a piece of furniture. 
\vspace{4mm}\\
\noindent \textbf{Example 6.}
\vspace{1mm}\\
\noindent cos\_sim$(chair) = [chairs, Chair, chairperson, $\\ 
$chairwoman, chairman, chairing, ...];$ 
\vspace{1mm}\\
\noindent $s$ = [$director, chairman, head, executive,$\\
$ president, speaker$];\\
$\alpha_1$ =  root$(s)$; $\alpha_2$ = $chair-\alpha_1;$\\
cos\_sim$(\alpha_2) = [chair, chairs, sofa, $\\$ couch, recliner, stool, chairwoman, ...];$\\

\subsubsection{Hypernymy}

The defined sub-vectors can be naturally considered the \emph{hypernyms}
of the children vectors as they represent the attribute which is shared by all of these word vectors.
Hence, the question arises, why are word vectors of hypernym words not sup-vectors of their hyponyms?
In fact, the cosine similarity between the word vectors of hyponyms and their hypernyms is often very small. 
We analyze this phenomenon in Example 7.

\vspace{4mm}
\noindent \textbf{Example 7.}
\vspace{1mm}\\
\noindent cos\_sim($months$) = [$month, week, August,$\\$ February, October, January, year, ..$]
\vspace{1mm}\\
\noindent $s$ = [$January, February, March, April, ...$];\\
$\alpha_1$ = root($s$); $|$ch($\alpha_1$)$|$  = 12  \\ 
$||\alpha_1||$ = 2.084; $||months||$ = 2.401;
\vspace{1mm}\\
\noindent cos\_sim($months-\alpha_1$) = [$months, weeks,$\\$ years, days, decades, quarters, ... $] \\

As the example illustrates, the word vector $months$ 
is also a member of a category which refers to temporal concepts, such as $months, weeks, and years$.
It follows that, hypernyms can also contain different meanings
which are not associated with their hyponyms.

\subsection{Semantic space networks} 
\label{sec:ssn}

\subsubsection{The semantic column model}

\begin{figure}
\def\x{0.4}
\def\y{0.875}
\def\z{1.0}
\def\a{\z}
\begin{tikzpicture}[scale=2]

    \draw [,] (0.0,0) -- (\z*2,0) ;
    
    \node[] at (\z - 0.075,0.2) {$\alpha$};
    \draw [thick,] (\z,0) -- (\z,\x-0.1);
    \node[] at (\z,\x) {$a$};
    \draw [thick,] (\z,\x) circle (0.1);

    \draw [thick,] (\z, 0.5) -- (\a, 0.75);

    \node[] at (\a,\y) {$v$};
    \node[] at (\a - 0.075,0.6) {$\beta$};
    \draw[thick,rotate around={45:(\a,\y)}] (\a-0.1,\y-0.1) rectangle (\a+0.1,\y+0.1);

\end{tikzpicture}
\centering
\caption{Semantic column: one single word vector is split into two sub-vectors}
\end{figure}
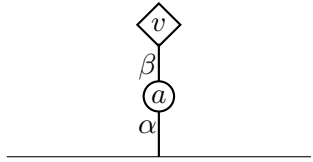

$v$ = $relation$

\noindent $\alpha$ = $0.6 \, v$;  \; par\_count($\alpha$) = 5;\\ 
ch($\alpha$) = 
[$relationship$, $relationships$, $\\$ $friendship$, $relation$, $ties$];\\

\noindent $\beta$ = $0.4 \, v $; \; par\_count($\beta$) = 17;\\ 
ch($\beta$) =  [ch($\alpha$), $cooperation,$ $romance,$ $alliance,$
$affair, rivalry, interaction, engagement, ...];$\\

\subsubsection{Ternary tree} 

\begin{figure}
\def\x{0.5}
\def\y{0.875}
\def\z{1.0}
\def\a{\z - 0.475}
\def\b{\z + 0.475}
\def\ya{1.25}
\def\za{0.52}
\def\aa{\za - 0.3}
\def\ba{1.0}
\def\yb{1.25}
\def\zb{1.48}
\def\ab{1}
\def\bb{\zb + 0.3}
\begin{tikzpicture}[scale=2]

    \draw [,] (0.0,0) -- (\z*2,0) ;
    
    \node[] at (\z,\x) {$a$};
    \node[] at (\z - 0.09,0.25) {$\alpha$};
    \draw [thick,] (\z,0) -- (\z,\x-0.1);
    \draw [thick,] (\z,\x) circle (0.1);

    \draw [thick,] (\z - 0.075, 0.575) -- (\a+0.1,0.8);
    \draw [thick,] (\z + 0.075, 0.575) -- (\b-0.1,0.8);

    \node[] at (\a,\y) {$b$};
    \node[] at (\a+0.2,0.6) {$\beta_1$};
    \draw [thick,] (\a,\y) circle (0.1);

    \node[] at (\b,\y) {$c$};
    \node[] at (\b-0.2,0.6) {$\beta_2$};
    \draw [thick,] (\b,\y) circle (0.1);

    \node[] at (\z, 0.75) {$O_1$};
    \draw [thick,] (\a+0.11,0.875) -- (\b-0.11,0.875);
    
    \draw [thick,] (\za - 0.075, \ya-0.3) -- (\aa+0.075,\ya-0.075);
    \draw [thick,] (\za + 0.075, \ya-0.3) -- (\ba-0.075,\ya-0.075);

    \node[] at (\aa,\ya) {$v_1$};
    \node[] at (\aa+0.6,\ya-0.25) {$\gamma_2$};
    \draw[thick,rotate around={45:(\aa,\ya)}] (\aa-0.1,\ya-0.1) rectangle (\aa+0.1,\ya+0.1);

    \node[] at (\ba,\ya) {$v_2$};
    \node[] at (\ba-0.7,\ya-0.25) {$\gamma_1$};
    \draw[thick,rotate around={45:(\ba,\ya)}] (\ba-0.1,\ya-0.1) rectangle (\ba+0.1,\ya+0.1);

    \node[] at (\za+0.1,\ya+0.125) {$O_2$};
    \draw [thick,] (\aa+0.125,\ya) -- (\ba-0.125,\ya);

    \draw [thick,] (\zb - 0.075, \yb-0.3) -- (\ab+0.075,\yb-0.075);
    \draw [thick,] (\zb + 0.075, \yb-0.3) -- (\bb-0.075,\yb-0.075);

    \node[] at (\ab+0.7,\yb-0.25) {$\gamma_4$};

    \node[] at (\bb,\yb) {$v_3$};
    \node[] at (\bb-0.6,\yb-0.25) {$\gamma_3$};
    \draw[thick,rotate around={45:(\bb,\yb)}] (\bb-0.1,\yb-0.1) rectangle (\bb+0.1,\yb+0.1);

    \node[] at (\zb-0.1,\yb+0.125) {$O_3$};
    \draw [thick,] (\ab+0.125,\yb) -- (\bb-0.125,\yb);

\end{tikzpicture}
\centering
\caption{Ternary tree: the interrelation of the three word vectors is explored by combining three semantic trees}
\end{figure}
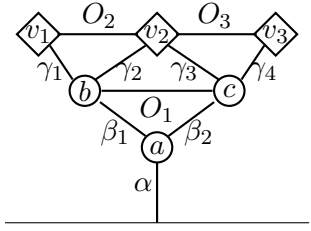

\noindent $s$ = $[v_1, v_2, v_3]$ = $[computer, mobile, camera];$

\noindent ch($\alpha$) = 
$[smartphone, handset, mobile,  $\\$ phones,  handsets, laptop,  camera, $\\$ browser, computer, desktop, BlackBerry];$\\

\noindent ch($\beta_1$) = 
$[software, Internet,  computer,  $\\$ IT,   internet, PC,  computers, Windows, virus, $\\$ Intel, database, machines, server, Data, $\\$ websites, electronics, ...];$\\
\noindent ch($\beta_2$) = 
$[photo, photos, camera, pictures, $\\$  shoot, cameras,  Photo, documentary,  footage, $\\$ photographer, filming, photography, portrait, $\\$ photograph, angle, filmed,  photographers];$\\

\noindent ch($\gamma_1$) = 
$[computer];$\\
ch($\gamma_2$) = 
$[mobile, Mobile];$\\
ch($\gamma_3$) = 
$[mobile];$\\
ch($\gamma_4$) = 
$[camera, cameras];$\\

\noindent ch($b$) = 
$[computer, smartphone, $\\$  computers, mobile];$\\
ch($c$) = 
$[camera, cameras, handset, $\\$ smartphone, phones, handsets, mobile];$\\

\subsubsection{Quadruple relations (analogy problem)}

\begin{figure}
\def\x{0.5}
\def\y{0.875}
\def\z{1.0}
\def\a{\z - 0.5}
\def\b{\z + 0.5}
\def\ya{1.25}
\def\za{0.52}
\def\aa{\za - 0.3}
\def\ba{\za + 0.3}
\def\yb{1.25}
\def\zb{1.48}
\def\ab{\zb - 0.3}
\def\bb{\zb + 0.3}
\begin{tikzpicture}[scale=2]

    \draw [,] (0.0,0) -- (\z*2,0) ;
    
    \node[] at (\z,\x) {$a$};
    \node[] at (\z - 0.09,0.25) {$\alpha_1$};
    \draw [thick,] (\z,0) -- (\z,\x-0.1);
    \draw [thick,] (\z,\x) circle (0.1);

    \draw [thick,] (\z - 0.075, 0.575) -- (\a+0.1,0.8);
    \draw [thick,] (\z + 0.075, 0.575) -- (\b-0.1,0.8);

    \node[] at (\a,\y) {$v_1$};
    \node[] at (\a+0.2,0.6) {$\beta_1$};
    \draw[thick,rotate around={0:(\a,\y)}] (\a-0.1,\y-0.1) rectangle (\a+0.1,\y+0.1);

    \node[] at (\b,\y) {$v_2$};
    \node[] at (\b-0.2,0.6) {$\beta_2$};
    \draw[thick,rotate around={0:(\b,\y)}] (\b-0.1,\y-0.1) rectangle (\b+0.1,\y+0.1);

    \node[] at (\z,0.98) {$O_1$};
    \draw [thick,] (\a+0.11,0.875) -- (\b-0.11, 0.875);

    \node[] at (\z,1.0+0.875+0.375) {$c$};
    \node[] at (\z - 0.1, 1.0 + 0.875 + 0.875 - 0.25) {$\alpha_3$};
    \draw [thick,] (\z,0.875+1+0.875) -- (\z,1.0+0.875+0.375+0.1);
    \draw [thick,] (\z,1.0+0.875+0.375) circle (0.1);

    \draw [thick,] (\z - 0.075,0.875 + 1 + 0.3) -- (\a+0.1,0.875+1+0.075);
    \draw [thick,] (\z + 0.075,0.875 + 1 + 0.3) -- (\b-0.1,0.875+1+0.075);

    \node[] at (\a,\y+1.0) {$v_4$};
    \node[] at (\a+0.2,0.275+1.0+0.875) {$\beta_6$};
    
    \draw[thick,rotate around={0:(\a,\y+1.0)}] (\a-0.1,\y-0.1+1.0) rectangle (\a+0.1,\y+0.1+1.0);

    \node[] at (\b,\y+1.0) {$v_3$};
    \node[] at (\b-0.2,0.275+1.0+0.875) {$\beta_5$};
    
    \draw[thick,rotate around={0:(\b,\y+1.0)}] (\b-0.1,\y-0.1+1.0) rectangle (\b+0.1,\y+0.1+1.0);

    \node[] at (\z, 1.0+0.750) {$O_3$};
    \draw [thick,] (\a+0.11, 1.0+0.875) -- (\b-0.11, 1.0+0.875);

    \draw [thick,] (\z - 0.5 - 0.375 - 0.1, 0.5+0.875) -- (\z - 0.5 - 0.875, 0.5+0.875);
    
    \node[] at (\z - 0.5 - 0.375 - 0.25,  0.5+0.875-0.1) {$\alpha_4$};
    
    \node[] at (\z - 0.5 - 0.375, 0.5+0.875) {$d$};
    \draw [thick,] (\z - 0.5 - 0.375, 0.5+0.875) circle (0.1);

    \draw [thick,] (\z - 0.5 - 0.3, 0.875 + 0.5 + 0.075) -- (\z - 0.5 - 0.075, 0.875 + 0.5 + 0.5 - 0.1);
    \draw [thick,] (\z - 0.5 - 0.3, 0.875 + 0.5 - 0.075) -- (\z - 0.5 - 0.075, 0.875 + 0.5 - 0.5 + 0.1);
    
    \node[] at (\z-0.5-0.275, 0.875+0.5-0.3) {$\beta_8$};
    \node[] at (\z-0.5-0.275, 0.875+0.5+0.3) {$\beta_7$};
   
    \node[] at (\z-0.35, 0.5+0.875) {$O_4$};
    \draw [thick,] (\z-0.5, 0.875+0.1) -- (\z-0.5, 1.0+0.875-0.1);

    \draw [thick,] (\z + 0.5 + 0.375 + 0.1, 0.5+0.875) -- (\z + 0.5 + 0.875, 0.5+0.875);
    
    \node[] at (\z + 0.5 + 0.375 + 0.25,  0.5+0.875-0.1) {$\alpha_2$};

    \node[] at (\z + 0.5 + 0.375, 0.5+0.875) {$b$};
    \draw [thick,] (\z + 0.5 + 0.375, 0.5+0.875) circle (0.1);

    \draw [thick,] (\z + 0.5 + 0.3, 0.875 + 0.5 + 0.075) -- (\z + 0.5 + 0.075, 0.875 + 0.5 + 0.5 - 0.1);
    \draw [thick,] (\z + 0.5 + 0.3, 0.875 + 0.5 - 0.075) -- (\z + 0.5 + 0.075, 0.875 + 0.5 - 0.5 + 0.1);
    
    \node[] at (\z+0.5+0.275, 0.875+0.5-0.3) {$\beta_3$};
    \node[] at (\z+0.5+0.275, 0.875+0.5+0.3) {$\beta_4$};
    
    \node[] at (\z+0.35, 0.5+0.875) {$O_2$};
    \draw [thick,] (\z+0.5, 0.875+0.1) -- (\z+0.5, 1.0+0.875-0.1);

     \draw [thick,] (\z + 0.875+0.5, 0.875 + 0.5 +0.2) -- (\z + 0.875+0.5,  0.875 + 0.5 -0.2);
     \draw [thick,] (\z - 0.875-0.5, 0.875 + 0.5 +0.2) -- (\z - 0.875-0.5,  0.875 + 0.5 -0.2);
     
     \draw [thick,] (\z - 0.2, 0.875 + 1 +0.875) -- (\z + 0.2, 0.875 + 1 +0.875);

\end{tikzpicture}
\centering
\caption{Quadruple relation: four semantic tree are joint}
\end{figure}
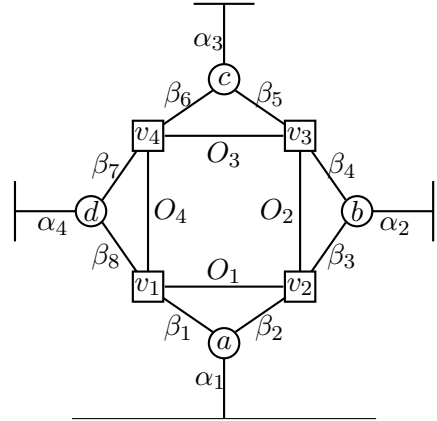

\noindent $[v_1, v_2, v_3, v_4]$ = $[Paris, France, Japan, Tokyo]$\\

\noindent ch($\alpha_1$) = 
$[Paris, France];$\\
ch($\alpha_2$) = 
$[Japan, France];$\\
ch($\alpha_3$) = 
$[Tokyo, Nikkei, yen, Japan];$\\
ch($\alpha_4$) = 
$[Tokyo, Paris];$\\

\noindent ch($\beta_1$) = 
$[Paris];$\\
ch($\beta_2$) = 
$[France];$\\
ch($\beta_3$) = 
$[France, Sarkozy];$\\
ch($\beta_4$) = 
$[Japan, yen, Tokyo, Nikkei];$\\

\noindent ch($\beta_5$) = 
$[Japan];$\\
ch($\beta_6$) = 
$[Tokyo, Shanghai, Seoul, Bangkok, $\\$ Manila, Frankfurt, Jakarta];$\\
ch($\beta_7$) = 
$[Japan, Japanese, yen, Tokyo, $\\$ Seoul, Pyongyang, Nikkei];$\\
ch($\beta_8$) = 
$[Paris]$\\

\end{document}